\documentclass[11pt,a4paper]{article}
\usepackage[hyperref]{emnlp-ijcnlp-2019}
\usepackage{times}
\usepackage{latexsym}
\usepackage{hyperref}
\usepackage{amsmath}
\usepackage{amssymb}
\usepackage{xcolor}
\usepackage{booktabs}
\usepackage{multirow}
\usepackage{soul}
\usepackage{url}
\usepackage{graphicx}
\usepackage{boldline}
\usepackage{makecell}
\usepackage{comment}

\aclfinalcopy % Uncomment this line for the final submission

\title{Domain Adaptive Text Style Transfer}

\author{\textbf{Dianqi Li}\textsuperscript{1},\hspace{2mm} \textbf{Yizhe Zhang}\textsuperscript{2},\hspace{2mm} \textbf{Zhe Gan}\textsuperscript{3},\hspace{2mm} \textbf{Yu Cheng}\textsuperscript{3}, \\
        \textbf{Chris Brockett}\textsuperscript{2},\hspace{2mm} \textbf{Ming-Ting Sun}\textsuperscript{1},\hspace{2mm} \textbf{Bill Dolan}\textsuperscript{2}\\
        \textsuperscript{1}University of Washington\hspace{2mm}
        \textsuperscript{2}Microsoft Research\hspace{2mm}
        \textsuperscript{3}Microsoft Dynamics 365 AI Research\\
        {\tt \small{\{dianqili,mts\}@uw.edu}}\\
        {\tt \small{ \{Yizhe.Zhang,Zhe.Gan,Yu.Cheng,Chris.Brockett,billdol\}@microsoft.com}}\\}

\date{}

\begin{document}
\maketitle
\begin{abstract}
Text style transfer without parallel data has achieved some practical success. However, in the scenario where less data is available, these methods may yield poor performance. In this paper, we examine domain adaptation for text style transfer to leverage massively available data from other domains. These data may demonstrate domain shift, which impedes the benefits of utilizing such data for training. To address this challenge, we propose simple yet effective domain adaptive text style transfer models, enabling domain-adaptive information exchange. The proposed models presumably learn from the source domain to: ($i$) distinguish stylized information and generic content information; ($ii$) maximally preserve content information; and ($iii$) adaptively transfer the styles in a domain-aware manner. We evaluate the proposed models on two style transfer tasks (sentiment and formality) over multiple target domains where only limited non-parallel data is available. Extensive experiments demonstrate the effectiveness of the proposed model compared to the baselines.
\end{abstract}

%---------------------------------------------------------------------
%Introduction
%---------------------------------------------------------------------
\section{Introduction}
Text style transfer, which aims to edit an input sentence with the desired style while preserving style-irrelevant content, has received increasing attention in recent years. It has been applied successfully to stylized image captioning~\cite{gan2017stylenet}, personalized conversational response generation~\cite{zhang2018personalizing}, formalized writing~\cite{rao2018dear}, offensive to non-offensive language transfer~\cite{dos2018fighting}, and other stylized text generation tasks \cite{akama2017stylecons,zhang2019consistent}.

Text style transfer has been explored as a sequence-to-sequence learning task using parallel datasets \cite{jhamtani2017shakespearizing}. However,
parallel datasets are often not available, and hand-annotating sentences in different styles is expensive. 
%Consequently, most previous text style transfer works consider a more realistic setting when only non-parallel stylized corpora are available. Such task is coined as unsupervised text style transfer \cite{yang2018unsupervised}. This makes the text style transfer even challenging because the style and content in the natural language are difficult to be disentangled without supervised signals from parallel data.
The recent surge of deep generative models~\cite{kingma2013auto,goodfellow2014generative} has spurred progress in text style transfer without parallel data by learning disentanglement~\cite{hu2017toward,shen2017style,fu2018style,li2018delete,prabhumoye2018style}. These methods typically require massive amounts of data~\cite{subramanian2018multiple}, and may perform poorly in limited data scenarios. 

A natural solution to the data-scarcity issue is to resort to massive data from other domains.
% It is often the case that only limited data is available in a target domain while a source domain contains plentiful data.
However, directly leveraging abundant data from other domains is problematic due to the discrepancies in data distribution on different domains. Different domains generally manifest themselves in domain-specific lexica. For example, sentiment adjectives such as  ``\emph{delicious}", ``\emph{tasty}", and ``\emph{disgusting}" in restaurant reviews might be out of place in movie reviews, where the sentiment words such as ``\emph{imaginative}", ``\emph{hilarious}", and ``\emph{dramatic}" are more typical.
%\dianqi{Obviously, a style-transferred sentence like ``the pizza is hilarious" is not suitable for restaurant reviews.}
Domain shift~\cite{domianshift} is thus apt to result in feature misalignment.
%, which prohibits directly leveraging massive available data from a source domain to a data-limited target domain.

% Previous works mainly consider the text style transfer within one domain distribution. 
In this work, we take up the problem of domain adaptation in scenarios where the target domain data is scarce and misaligned with the distribution in the source domain. Our goal is to achieve successful style transfer into the target domain, with the help of the source domain, while the transferred sentences carry relevant characteristics in the target domain. 

We present two first-of-their-kind domain adaptive text style transfer models that facilitate domain-adaptive information exchange between the source and target domains. These models effectively learn generic content information and distinguish domain-specific information. Generic content information, primarily captured by modeling a large corpus from the source domain, facilitates better content preservation on the target domain. Meanwhile, domain-specific information, implicitly imposed by domain vectors and domain-specific style classifiers, underpins the transferred sentences by generating target-specific lexical terms.
%Extensive experiments show that the proposed model attains favorable performance on two style transfer tasks (sentiment and formality) over multiple data-sparse target domains. 

Our contributions in this paper are threefold: ($i$) We explore a challenging domain adaptation problem for text style transfer by leveraging massively-available data from other domains.  ($ii$) We introduce simple text style transfer models that preserve content and meanwhile translate text adaptively into target-domain-specific terms. ($iii$) We demonstrate through extensive experiments the robustness of these methods for style transfer tasks (sentiment and formality) on multiple target domains where only limited non-parallel data is available. Our implementation is available at \url{https://github.com/cookielee77/DAST}.

%-------------------------------
%Related work
%-------------------------------
\section{Related Work}

\paragraph{Text Style Transfer.} Text style transfer using neural networks has been widely studied in the past few years. A common paradigm is to first disentangle latent space as content and style features, and then generate stylistic sentences by tweaking the style-relevant features and passing through a decoder. \citet{hu2017toward,fu2018style,shen2017style,yang2018unsupervised,gong2019reinforcement,lin2017adversarial} explored this direction by assuming the disentanglement can be achieved in an auto-encoding procedure with a suitable style regularization, implemented by either adversarial discriminators or style classifiers. \citet{li2018delete,xu2018unpaired,zhang2018learning} achieved disentanglement by filtering the stylistic words of input sentences. Recently, \citet{prabhumoye2018style} has proposed to use back-translation for text style transfer with a de-noising auto-encoding objective~\cite{logeswaran2018content,subramanian2018multiple}. Our work differs from the above in that we leverage domain adaptation to deal with limited target domain data, whereas previous methods require massive target domain style-labelled samples.

% Existing works mainly aim at transferring the style of texts by using a massive amount of non-parallel stylistic samples. In contrast, our work presents two general learning frameworks together with domain adaptation to solve the problem when the target style transfer task only has a small amount of data. 

\paragraph{Domain Adaptation.}
Domain adaptation has been studied in various natural language processing tasks, such as sentiment classification~\cite{qu2019adversarial}, dialogue system~\cite{wen2016multi}, abstractive summarization~\cite{hua2017pilot,zhang2018shaped}, machine translation~\cite{koehn2007experiments,axelrod2011domain,sennrich2016improving,michel2018extreme}, etc. However, no work has been done for exploring domain adaptation on text style transfer. To our best knowledge, we are the first to explore the adaptation of text style transfer models for a new domain with limited non-parallel data available. The task requires both style transfer and domain-specific generation on the target domain. To differentiate different domains,  \citet{sennrich2016controlling,chu2017empirical} appended domain tokens to the input sentences.  Our model uses learnable domain vectors combining domain-specific style classifiers, which force the model to learn distinct stylized information in each domain.

% To mitigate the effects of domain shift, there have been many efforts that study domain adaption in different NLG tasks, such as dialogue system~\cite{wen2016multi}, abstractive summarization~\cite{hua2017pilot,zhang2018shaped}, machine translation~\cite{koehn2007experiments,axelrod2011domain,sennrich2016improving,chu2017empirical}, etc. \dq{May be explain more about domain adaptation method? e.g., feature alignment, mix-training and fine-tuning training.} However, few work has been done for exploring domain adaption of text style transfer without parallel data. To our best knowledge, we are the first to study the adaptation of text style transfer models for a new domain with limited non-parallel data available.

%-------------------------------
%Preliminary
%-------------------------------
\section{Preliminary}
We first describe a standard text style transfer approach, which only considers data in the target domain. We limit our discussion to the scenario where only non-parallel data is available, since large amounts of parallel data is typically not feasible. 

Given a set of style-labelled sentences $\mathcal{T} = \{(x_i, l_i)\}^{N}_{i = 1}$ in the target domain, the goal is to transfer sentence $x_i$ with style $l_i$ to a sentence $\widetilde{x}_i$ with another style $\widetilde{l}_i$, where $\widetilde{l}_i \neq l_i$. $l_i, \widetilde{l}_i$ belong to a set of style labels $l^{\mathcal{T}}$ in the target domain: $l_i, \widetilde{l}_i \in l^{\mathcal{T}}$. Typically, an encoder encodes the input $x_i$ to a semantic representation $c_i$, while a decoder controls or modifies the stylistic property and decodes the sentence $\widetilde{x}_i$ based on $c_i$ and the pre-specific style $\widetilde{l}_i$. 

Specifically, we denote an encoder-decoder model as $(E, D)$. The semantic representation $c_i$ of sentence $x_i$ is extracted by the encoder $E$, i.e., $c_i = E(x_i)$. The decoder $D$ aims to learn a conditional distribution of $\widetilde{x_i}$ given the semantic representation $c_i$ and style $\widetilde{l_i}$:
\begin{equation}
    p_D(\widetilde{x}_i | c_i, \widetilde{l}_i) = \prod^{T}_{t=1} p_D(\widetilde{x}_i^t | \widetilde{x}_i^{<t}, c_i, \widetilde{l}_i),
    \label{eq:1}
\end{equation}
where $\widetilde{x}_i^t$ is the $t^{th}$ token of $\widetilde{x}_i$, and $\widetilde{x}_i^{<t}$ is the prefix of $\widetilde{x}_i$ up to the $t^{th}$ token.

Directly estimating Eqn.~\eqref{eq:1} is impractical during training due to a lack of parallel data $(x_i, \widetilde{x}_i)$. Alternatively, the original sentence $x_i$ should have high probability under the conditional distribution $p_D(x_i | c_i, l_i)$. Thus, an auto-encoding reconstruction loss could be formulated as:
\begin{equation}
    L_{ae}^{\mathcal{T}} = -\mathop{\mathbb{E}}_{x_i \sim \mathcal{T}} \text{log } p_D(x_i | c_i, l_i)\,.
    \label{eq:2}
\end{equation}
Note that we assume that the decoder $D$ recovers $x_i$'s original stylistic property as accurate as possible when given the style label $l_i$. To achieve text style transfer, the decoder manipulates the style of generated sentences by replacing $l_i$ with a desired style $\widetilde{l}_i$. Specifically, the generated sentence $\widetilde{x}_i$ is sampled from $\widetilde{x}_i \sim p_D(\widetilde{x}_i | c_i, \widetilde{l}_i)$. However, by directly optimizing Eqn.~\eqref{eq:2}, the encoder-decoder model tends to ignoring the style labels and collapses to a reconstruction model, which might simply copy the input sentence, hence fails to transfer the style. To force the model to learn meaningful style properties, \citet{hu2017toward,hu2018texar} apply a style classifier for the style regularization. The style classifier ensures the encoder-decoder model to transfer $\widetilde{x}_i$ with its correct style label $\widetilde{l}_i$:
\begin{equation}
    L_{style}^\mathcal{T} = -\mathop{\mathbb{E}}_{\widetilde{x}_i \sim p_D(\widetilde{x}_i | c_i, \widetilde{l}_i)} \text{log } P_{C^\mathcal{T}}(\widetilde{l}_i | \widetilde{x}_i)\,,
    \label{eq:3}
\end{equation}
where $C^\mathcal{T}$ is the style classifier pretrained on the target domain. The overall training objective for text style transfer within the target domain $\mathcal{T}$ is written as: 
\begin{equation}
    L^\mathcal{T} = L_{ae}^\mathcal{T} + L_{style}^\mathcal{T}\,.
    \label{eq:4}
\end{equation}

%-------------------------------
%Method
%-------------------------------
\section{Domain Adaptive Text Style Transfer}
In this section, we present Domain Adaptive Style Transfer (DAST) models to perform style transfer on a target domain by borrowing the strength from a source domain, while maintaining the transfer to be domain-specific. 

\subsection{Problem Definition}
\label{sec:3.1}

Suppose we have two sets of style-labelled sentences $\mathcal{S} = \{(x^\prime_i, l^\prime_i)\}^{N^\prime}_{i = 1}$, $\mathcal{T} = \{(x_i, l_i)\}^{N}_{i = 1}$ in the source domain $\mathcal{S}$ and the target domain $\mathcal{T}$, respectively. $x^\prime_i$ denotes the $i^{th}$ source sentence. $l^\prime_i$ denotes the corresponding style label, which belongs to a source style label set: $l^\prime_i \in l^\mathcal{S}$ (e.g., positive/negative). $l^\prime_i$ can be available or unknown. Likewise, pair $(x_i, l_i)$ represents the sentence and style label in the target domain, where $l_i \in l^\mathcal{T}$. 

We consider domain adaptation in two settings: $(i)$ the source style $l^\mathcal{S}$ is unknown, 
%to the target style, $l^\mathcal{T} \neq l^\mathcal{S}$,
e.g., we may have a large corpus, such as Yahoo! Answers, but the underlying style for each sample is not available;
%in the same style labels of positive/negative sentiments as Yelp restaurant reviews. 
$(ii)$ the source styles are available, and are the same as the target styles, i.e., $l^\mathcal{T} = l^\mathcal{S}$, e.g., both IMDB movie reviews and Yelp restaurant reviews have the same style classes (negative and positive sentiments).

In both scenarios, we assume that the target domain $\mathcal{T}$ only has limited non-parallel data. With the help of source domain data $\mathcal{S}$, the goal is to transfer $(x_i, l_i)$ to $(\widetilde{x}_i, \widetilde{l}_i)$ in the target domain. The transferred sentence $\widetilde{x}_i$ should simultaneously hold: $(i)$ the main content with $x_i$, $(ii)$ a different style $\widetilde{l}_i$ from $l_i$, and $(iii)$ domain-specific characteristics of the target data distribution $\mathcal{T}$.

\begin{figure*}[t]
%\minipage{0.49\textwidth}
  \includegraphics[width=0.49\linewidth]{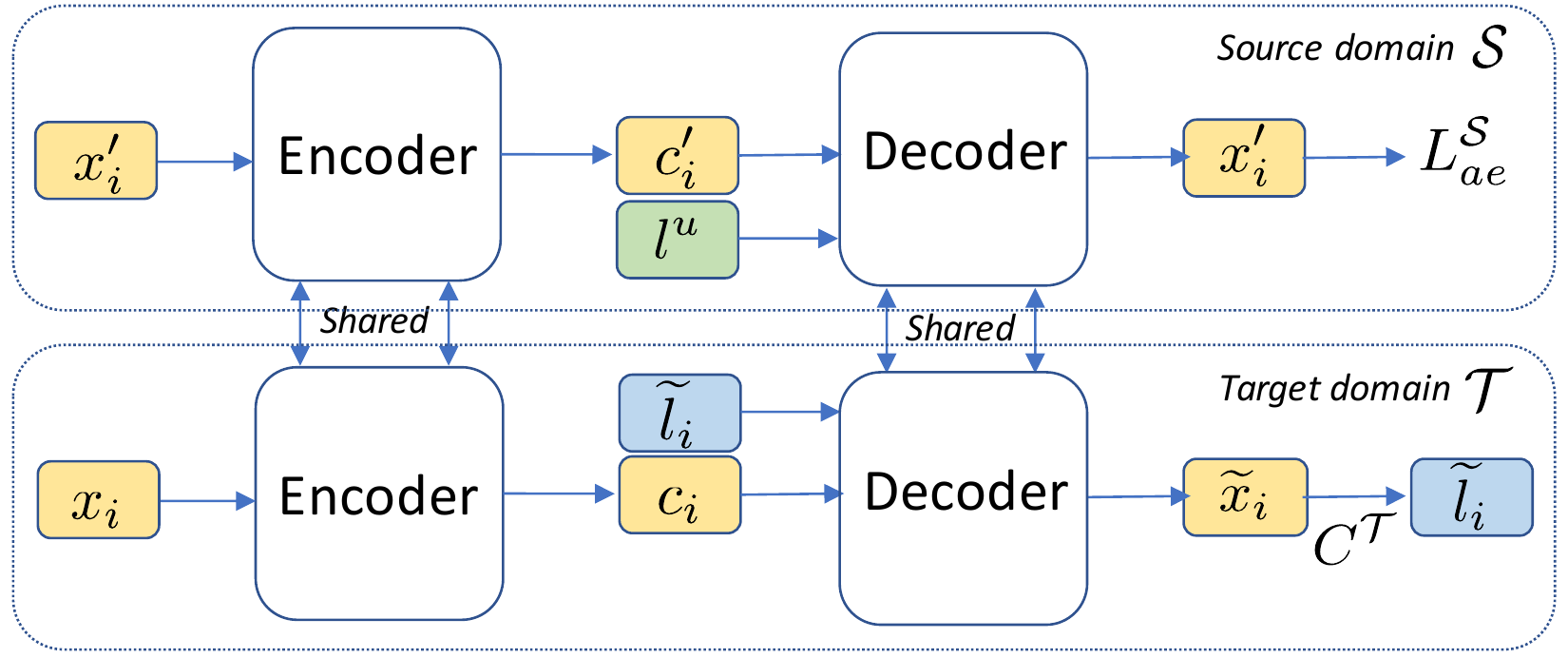}
  \includegraphics[width=0.49\linewidth]{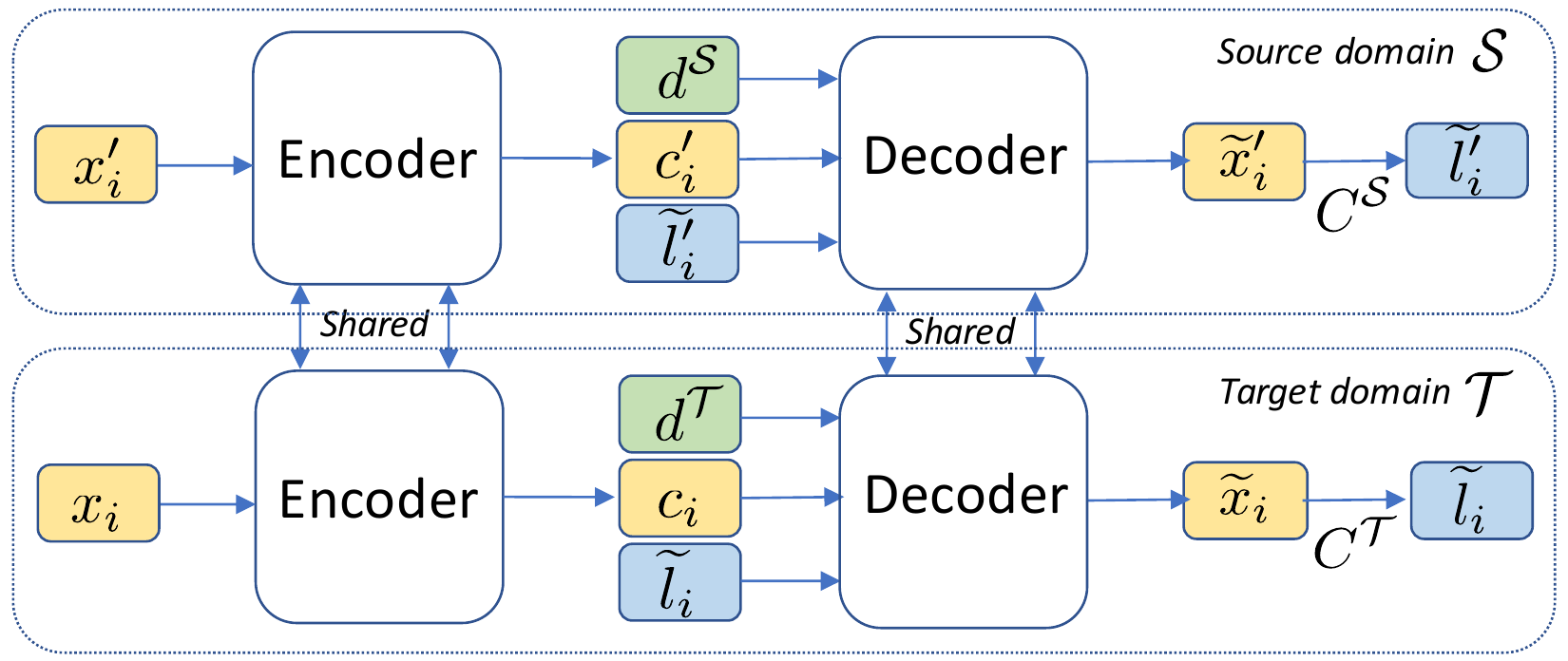}
  \caption{Illustration of the proposed \textbf{DAST-C} (left) and \textbf{DAST} (right) model.  DAST-C learns the generic content information through $L^\mathcal{S}_{ae}$ on massive source domain data with unknown style $l^u$.  For DAST, $d^\mathcal{T}, d^\mathcal{S}$ and $C^\mathcal{T}, C^\mathcal{S}$ denote domain vectors and domain-specific style classifiers, respectively. Better looked in color.}
  \label{fig:DAST}
\end{figure*}

\subsection{DAST with unknown-stylized source data}
\label{sec:3.3}
In this section, we investigate the case that the source style $l^\mathcal{S}$ is unknown. We first examine a drawback of limited target data to motivate our method. With limited target data, Eqn.~\eqref{eq:4} may yield an undesirable transferred text, where the generated text tends to using the most discriminative words that the target style prefers while ignoring the content. This is because the classifier $C^\mathcal{T}$ typically requires less data to train, comparing with a sequence autoencoder $(E, D)$. The classifier objective $L_{style}^\mathcal{T}$ thus dominates Eqn.~\eqref{eq:4}, rendering the generator to bias the sentences with most representative stylized (e.g., positive or negative) words rather than preserving the contents (see Table~\ref{tab:quali} for examples).

We consider alleviating this issue by leveraging massive source domain data to enhance the content-preserving ability, though the underlying styles in the source domain are unknown. By jointly training an auto-encoder on both the source and target domain data, the learned generic content information enables the model to yield better content preservation on the target domain.

To utilize the source data, we consider that $l^{\mathcal{S}}$ only contains a special unknown-style label $l^u$, separated from the target style $l^\mathcal{T}$. We assume the semantic representation of the source data $c^\prime_i$ is encoded by the encoder, i.e., $c^\prime_i = E(x^\prime_i)$. The decoder takes $c^\prime_i$ with style $l^u$ to generate the sentences on the source domain. The auto-encoding reconstruction objective of the source domain is:
\begin{equation}
    L_{ae}^{\mathcal{S}} = -\mathop{\mathbb{E}}_{x^\prime_i \sim \mathcal{S}} \text{log } p_D(x^\prime_i | c^\prime_i, l^u),
\end{equation}
where the encoder-decoder model $(E, D)$ is shared in both domains. Therefore, the corresponding objective can be written as:
\begin{equation}
    L_{\text{DAST-C}} = L_{ae}^\mathcal{T} + L_{style}^\mathcal{T} + L_{ae}^{\mathcal{S}}\,.
    \label{eq:6}
\end{equation}
This can be perceived as combining the source domain data with the target domain data to train a better encoder-decoder framework, while target-specific style information on the target domain is learned through $L_{style}^\mathcal{T}$.

Note that $L_{ae}^\mathcal{T}$ and $L_{ae}^{\mathcal{S}}$ are conditional on domain-specific styles labels: $l^\mathcal{T}$ and $l^u$, which implicitly encourages the model to learn domain-specific features. The decoder could thus generate target sentences adaptively with $l^\mathcal{T}$, while achieving favorable content preservation with the generic content information modeled by $L_{ae}^{\mathcal{S}}$. We refer this model, which is illustrated in Figure~\ref{fig:DAST}(left), as \textit{Domain Adaptive Style Transfer with generic Content preservation (DAST-C)}.

\subsection{DAST with stylized source data}
We further explore the scenario where $l^\mathcal{S} = l^\mathcal{T}$. In this case, besides the generic content information, there is much style information from the source domain that could be leveraged, e.g., generic stylized expressions like ``\emph{fantastic}'' and ``\emph{terrible}'' for sentiment transfer can be applied to both restaurant and movie reviews. We thus consider to borrow the full strength of the source data, by sharing learned knowledge on both the generic content and style information.

A straightforward way to achieve this is to train Eqn.~\eqref{eq:4} on both domains. However, simply mixing the two domains together will lead to undesirable style transfers, where the transfer is not domain-specific. For example, when adapting the IMDB movie reviews to the Yelp restaurant reviews, directly sharing the style transfer model without specifying the domain will inevitably result in generations like \textit{``The pizza is dramatic!''}.

To alleviate this problem, we introduce additional domain vectors, encouraging the model to perform style transfer in a domain-aware manner. The proposed DAST model is illustrated in Figure~\ref{fig:DAST}(right). Consider two domain vectors: $d^\mathcal{S}$ for the source domain and $d^\mathcal{T}$ for the target domain, respectively. We rewrite the auto-encoding loss as:
\begin{equation}
\begin{aligned}
    L_{ae}^{\mathcal{S,T}} = &-\mathop{\mathbb{E}}_{x^\prime_i \sim \mathcal{S}} \text{log } p_D(x^\prime_i | c^\prime_i, d^\mathcal{S}, l^\prime_i) \\
    &-\mathop{\mathbb{E}}_{x_i \sim \mathcal{T}} \text{log } p_D(x_i | c_i, d^\mathcal{T}, l_i)\,,
\end{aligned}
\end{equation}
where the encoder-decoder model $(E, D)$ is shared across domains. The domain vectors, $d^\mathcal{S}$, $d^\mathcal{T}$, learned from the model, implicitly guide the decoder to generate sentences with domain-specific characteristics. Note that $l_i$ and $l^\prime_i$ are shared, i.e., $l^\mathcal{T}=l^\mathcal{S}$. This enables the model to learn generic style information from both domains. On the other hand, explicitly learning precise stylized information within each domain is crucial to generate domain-specific styles. Thus, two domain-specific style classifiers ensure the model to learn the corresponding styles by conditioning on $(d^\mathcal{S}, \widetilde{l}^\prime_i)$ in the source domain or $(d^\mathcal{T}, \widetilde{l}_i)$ in the target domain:
\begin{equation}
\begin{aligned}
    L_{style}^\mathcal{S, T} = &-\mathop{\mathbb{E}}_{{\scriptstyle\widetilde{x}^\prime_i \sim p_D(\widetilde{x}^\prime_i | c^\prime_i, d^\mathcal{S}, \widetilde{l}^\prime_i)}} \text{log } P_{C^\mathcal{S}}(\widetilde{l}^\prime_i | \widetilde{x}^\prime_i) \\
    &-\mathop{\mathbb{E}}_{{\scriptstyle\widetilde{x}_i \sim p_D(\widetilde{x}_i | c_i, d^\mathcal{T}, \widetilde{l}_i)}} \text{log } P_{C^\mathcal{T}}(\widetilde{l}_i | \widetilde{x}_i)\,,
\end{aligned}
\end{equation}
where $\widetilde{x}^\prime_i, \widetilde{x}_i$ are the transferred sentences with pre-specific styles $\widetilde{l}^\prime_i, \widetilde{l}_i$ in the source and target domains, respectively. The domain-specific style classifiers, $C^\mathcal{T}$ and $C^\mathcal{S}$, are trained separately on each domain. The signals from classifiers encourage the model to learn domain-specific styles combining with the domain vectors and style labels. The overall training objective of the proposed DAST model is:
\begin{equation}
    L_{\text{DAST}} = L_{ae}^\mathcal{S,T} + L_{style}^\mathcal{S,T}\,.
    \label{eq:9}
\end{equation}
The domain-specific style classifiers enforce the model to learn domain-specific style information conditioning on $(d^\mathcal{S}, \widetilde{l}^\prime_i)$ or $(d^\mathcal{T}, \widetilde{l}_i)$, which in turn controls the model to generate sentences with domain-specific words. The model can thus distinguish domain-specific features, and adaptively transfer the styles in a domain-aware manner.

%Since the style information in both domain is similar, a straightforward way to adapt the source-trained style transfer model to the target domain is to continue fine-tuning on the target samples. However, we empirically observe that over fine-tuning will eliminate beneficial language and style information captured on the source domain. Due to a lack of obvious training signals in unsupervised text style transfer learning, the style transfer model will dramatically overfit the objectives on the target domain, resulting in a catastrophic forgetting problem~\cite{goodfellow2013empirical}. 

%---------------------------------------------------------------------
%Experiments
%---------------------------------------------------------------------
\section{Experiments}
We evaluate our proposed models on two tasks: sentiment transfer (positive-to-negative and negative-to-positive), and formality transfer (informal-to-formal). In both tasks, we make comparisons with previous approaches over multiple target domains. All experiments are conducted on one Nvidia GTX 1080Ti GPU.%\footnote{Our code and data will be released upon acceptance.}. 

\subsection{Dataset}

A statistics for the source and target corpora used in the experiments is summarized in Table~\ref{tab:1}.

\begin{table}[htbp]
\centering
\small{
\begin{tabular}{c|c|c|c|c|c}
\hline
\multicolumn{6}{c}{Sentiment Transfer}\\\hline
Source   & Train   & Target    & Train   & Dev & Test\\\hline
\multirow{3}{*}{\textsc{IMDB}} & \multirow{3}{*}{344k} & \textsc{Yelp} & 444k & 4k & 1k \\\cline{3-6}
 & & \textsc{Amazon} & 554k & 2k & 1k \\\cline{3-6}
  & & \textsc{Yahoo} & 4k & 2k & 1k \\\hline
  
  \multicolumn{6}{c}{Formality Transfer}\\\hline
  Source   & Train   & Target    & Train   & Dev & Test\\\hline
  \textsc{GYAFC} & 103k & \textsc{Enron} & 6k & 500 & 500 \\\hline
\end{tabular}
}
\caption{Statistics of source and target datasets.}
\vspace{-4mm}
\label{tab:1}
\end{table}

\paragraph{Sentiment Transfer.} 
For the source domain, we use IMDB movie review corpus~\cite{diao2014jointly} by following the filtering and preprocessing pipelines from~\citet{shen2017style}. This results in $344k$ training samples with sentiment labels. For the target domain, both the Yelp restaurant review dataset and the Amazon product review dataset are from~\citet{li2018delete}. For the test sets, we evaluate our methods by using $1k$ human-transferred sentences, annotated by~\citet{li2018delete}, on both Yelp and Amazon datasets. In addition to the two standard sentiment datasets, we manually collected a Yahoo sentimental question dataset - $7k$ question samples with sentiments from Yahoo! Answers dataset~\cite{zhang2015character}. We split the $7k$ sentimental questions into $4k$/$2k$/$1k$ for train/dev/test sets, respectively. Note that the Yahoo sentiment dataset only consists of questions, which have different domain characteristics with the IMDB dataset. In all the sentiment experiments, we consider both transfer directions (positive-to-negative and negative-to-positive).

\paragraph{Formality Transfer.}
We use the Grammarly's Yahoo Answers Formality Corpus (GYAFC)~\cite{rao2018dear} as the source dataset. The publicly released version of GYAFC only covers two topics  (\textit{Entertainment \& Music} and \textit{Family \& Relationships}), where each topic contains $50k$ paired informal and formal sentences written by humans. For the target domain, we use Enron email conversation dataset\footnote{https://www.cs.cmu.edu/\textasciitilde./Enron/}, which covers several different fields like \textit{Business, Politics, Daily Life, etc.} We manually labeled $7k$ \textit{non-parallel} sentences written in either the formal or informal style. We split the Enron dataset into $6k, 500, 500$ samples for training, validation and testing, respectively. Both the validation and test set consist of mere informal sentences, where the corresponding formal references are annotated by us from a crowd-sourcing platform for evaluation. We only assess the informal-to-formal transfer direction in the formality transfer experiment. 

\begin{table*}[t]
\centering
\small{
\begin{tabular}{ccccc|cccc}
\toprule
{} & \multicolumn{4}{c}{Yelp} & \multicolumn{4}{c}{Amazon}\\\midrule
Model (100\% target data) &  D-acc & S-acc   & \textit{h}BLEU    & G-score   & D-acc & S-acc   & \textit{h}BLEU    & G-score\\\midrule
CrossAlign~\cite{shen2017style}   &- & 85.0  & 3.7   & 8.3   &- & 23.0   & 34.1    & 18.0\\
Delete\&Retrieve~\cite{li2018delete} &- &90.6 &14.8 &17.9 &- &50.9 &30.3 &25.7\\
CycleRL~\cite{xu2018unpaired} &- &88.7 &12.3 &16.4 &- &68.7 &14.2 &15.5\\
SMAE~\cite{zhang2018learning} &- &85.1 &12.1 &15.5 &- &71.1 &12.9 &14.9\\
ControlGen~\cite{hu2018texar} & &91.5 &25.5 &27.4 &- &79.0 &31.1 & 30.5\\
Finetune &\textbf{96.1} &91.3 &25.6 &27.8 &\textbf{97.4} &79.2 &34.1 & 34.3\\
DAST-C (ours) &93.8 &91.7 &25.7 &27.5 &96.7 &81.9 &35.7 &35.0\\
DAST (ours) &95.8 &\textbf{92.3} &\textbf{26.3} &\textbf{28.9} &96.9 &\textbf{83.0} &\textbf{35.9} &\textbf{35.1}\\\midrule
Model (1\% target data) &  D-acc & S-acc   & \textit{h}BLEU    & G-score   & D-acc & S-acc   & \textit{h}BLEU    & G-score\\\midrule
CrossAlign~\cite{shen2017style} &- &76.3 &4.8 &8.5 &- &83.2 &2.0 &5.9\\
Delete\&Retrieve~\cite{li2018delete} &- &82.1 &4.1 &7.6 &- &63.0 &6.9 &9.3\\
CycleRL~\cite{xu2018unpaired} &- &86.6 &1.4 &5.2 &- &79.5 &0.7 &3.8\\
SMAE~\cite{zhang2018learning} &- &96.0 &1.2 &4.8 &- &87.2 &0.4 &3.2\\
ControlGen~\cite{hu2018texar} &- &\textbf{98.5} &3.7 &8.6 &- &83.2 &1.9 &5.8\\
Finetune &\textbf{98.1} &96.7 &13.9 &18.5 &\textbf{96.0} &\textbf{89.2} &11.3 &14.4\\
DAST-C (ours) &96.9 &90.3 &17.8 &19.3 &94.8 &78.2 &20.1 &21.6\\
DAST (ours) &97.0 &92.6 &\textbf{20.1} &\textbf{23.1} &94.6 &82.7 &\textbf{21.0} &\textbf{23.1}\\\bottomrule
\end{tabular}
}
\caption{Automatic evaluation results on Yelp and Amazon test sets. D-acc and S-acc denote domain accuracy and style accuracy, respectively. G-score is the geometric mean of S-acc and \emph{h}BLEU.}
\label{tab:res}
\end{table*}

\subsection{Evaluation}
%We evaluate our proposed frameworks on both automatic and human evaluations.
\paragraph{Automatic Metrics. }
We evaluate the effectiveness of our DAST models based on three automatic metrics:  

\noindent ($i$) Content Preservation. We assess the content preservation according to n-gram statistics, by measuring the BLEU scores~\cite{papineni2002BLEU} between generated sentences and human references on the target domain, refered as \textit{human} BLEU (\textit{h}BLEU). When no human reference is available (e.g., Yahoo), we compute the BLEU scores with respect to the input sentences. 

\noindent ($ii$) Style Control. We generate samples from the model and measure the style accuracy with a style classifier that is pre-trained on the target domain. We refer the style accuracy as S-acc. 

\noindent ($iii$) Domain Control. To validate whether the generated sentences hold the characteristics of the target domain, we adopt a pre-trained domain classifier to measure the percentage of generated sentences that belong to the target domain. We refer the domain accuracy as D-acc. 

All the pre-trained classifiers are implemented by TextCNN ~\cite{kim2014convolutional, zhang2017deconvolutional}. The test accuracy of all these classifiers used for evaluation are reported in Appendix~\ref{supp:classifers}. Following~\citet{xu2018unpaired}, we also evaluate all methods using a single unified metric called G-score, which calculates the geometric mean of style accuracy and \textit{h}BLEU.

\paragraph{Human Evaluation.}
To accurately evaluate the quality of transferred sentences, we conduct human evaluations based on the \textit{content preservation}, \textit{style control} and \textit{fluency} aspects by following~\citet{mir2019evaluating}. Previous works~\cite{subramanian2018multiple,gong2019reinforcement} ask workers to evaluate the quality via a numerical score, however, we found that this empirically leads to high-variance results. Instead, we pair transferred sentences from two different models, and ask workers to choose the sentence they prefer \emph{when compared to the input} on each evaluation aspect. We provide a ``No Preference" option to choose when the workers think the qualities of the two sentences are indistinguishable. Details of the human evaluation instruction are included in Appendix~\ref{supp:human_eval}. For each testing, we randomly sample 100 sentences from the corresponding test set and collect three human responses for each pair on every evaluation aspect, resulting in 2700 responses in total.

\subsection{Experimental Setup}

The encoder $E$ and the decoder $D$ are implemented by one-layer GRU~\cite{cho2014learning} with hidden dimensions 500 and 700, respectively. The domain-vector dimension is set to 50. The style labels are represented by learnable vectors with 150 dimensions. The decoder is initialized by a concatenation of representations of content, style, and domain vectors. If domain vectors are not used, the dimension of style labels is set to 200; accordingly, the initialization of the decoder is a concatenation of content and style representations.  TextCNN~\cite{kim2014convolutional} is employed for the domain-specific style classifiers pre-trained on corresponding domains. After pre-training, the parameters of the classifiers are fixed. We use the hard-sampling trick~\cite{logeswaran2018content} to back-propagate the loss through discrete tokens from the classifier to the encoder-decoder model. During training, we assign each mini-batch the same amount of source and target data to balance the training.

We make an extensive comparison with five state-of-the-art text style transfer models: CrossAlign~\cite{shen2017style}, Delete\&Retrieve~\cite{li2018delete}, CycleRL~\cite{xu2018unpaired}, SMAE~\cite{zhang2018learning} and ControlGen~\cite{hu2018texar}. We also experiment a simple and effective domain adaptation baseline - Finetune, which is trained with Eqn.~\eqref{eq:4} on the source domain and then fine-tuned on the target domain.

\subsection{Results}

\begin{figure*}[htbp]
\minipage{0.25\textwidth}
  \includegraphics[width=\linewidth]{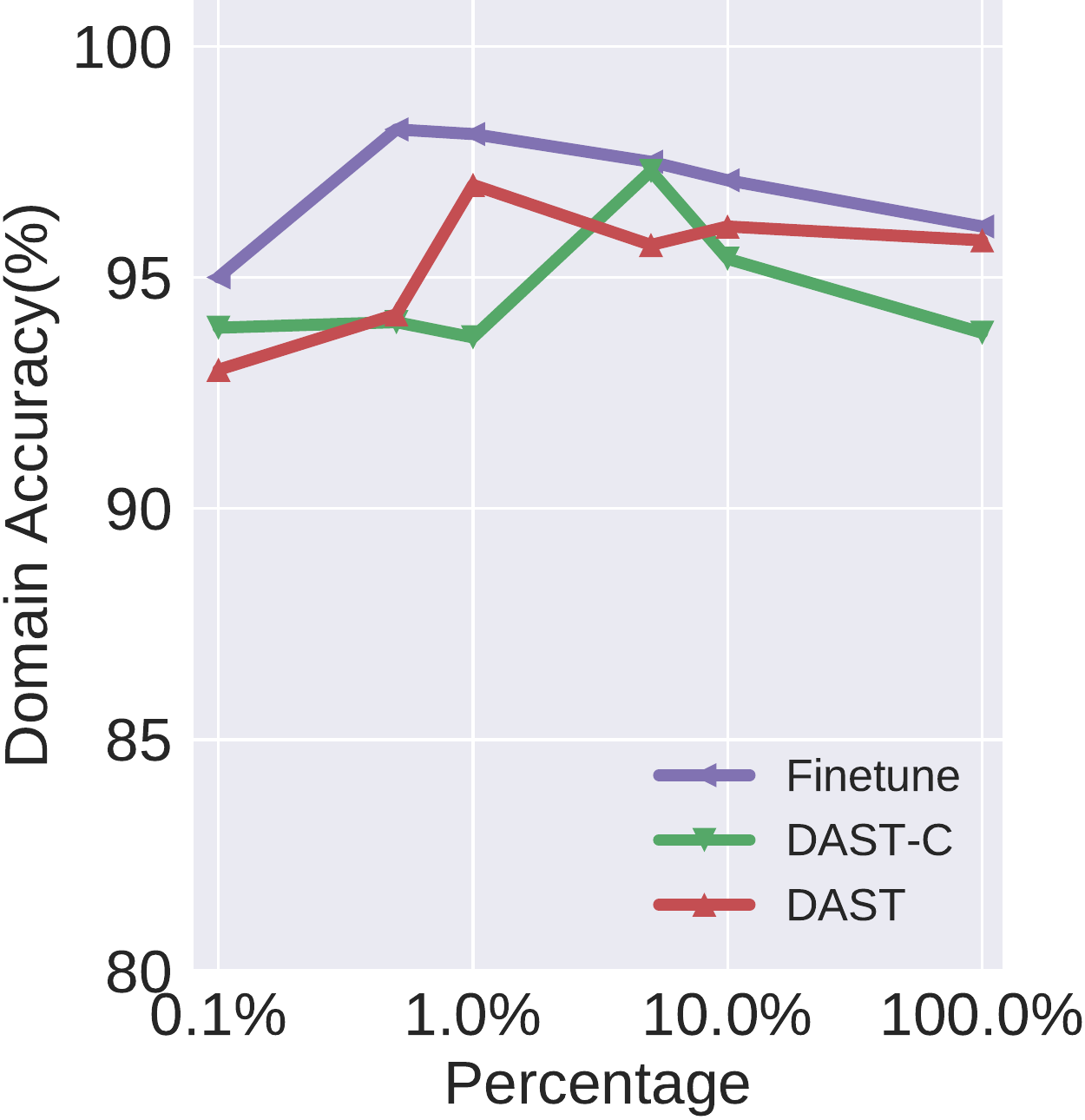}
\endminipage\hfill
\minipage{0.25\textwidth}
  \includegraphics[width=\linewidth]{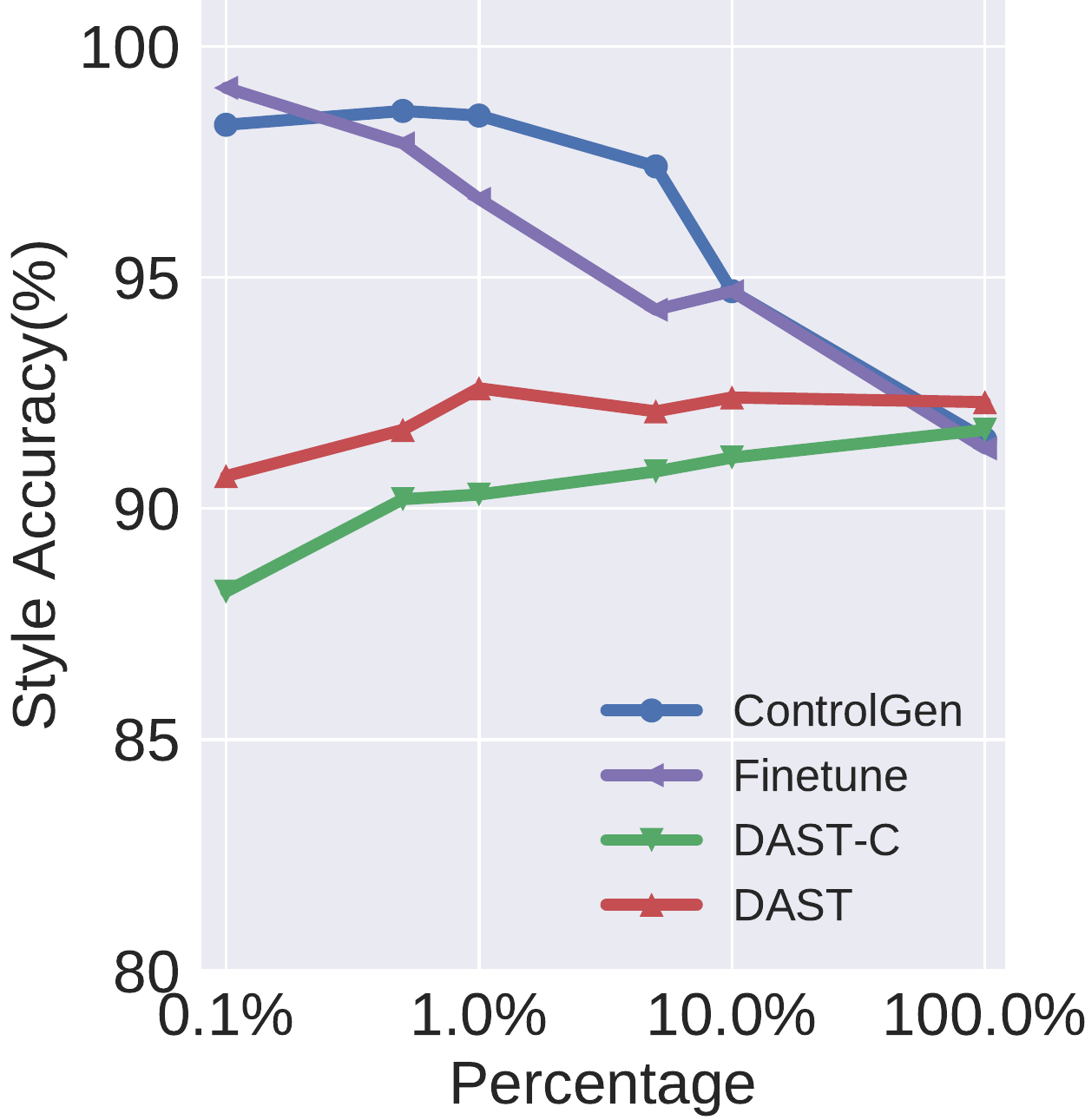}
\endminipage\hfill
\minipage{0.25\textwidth}
  \includegraphics[width=\linewidth]{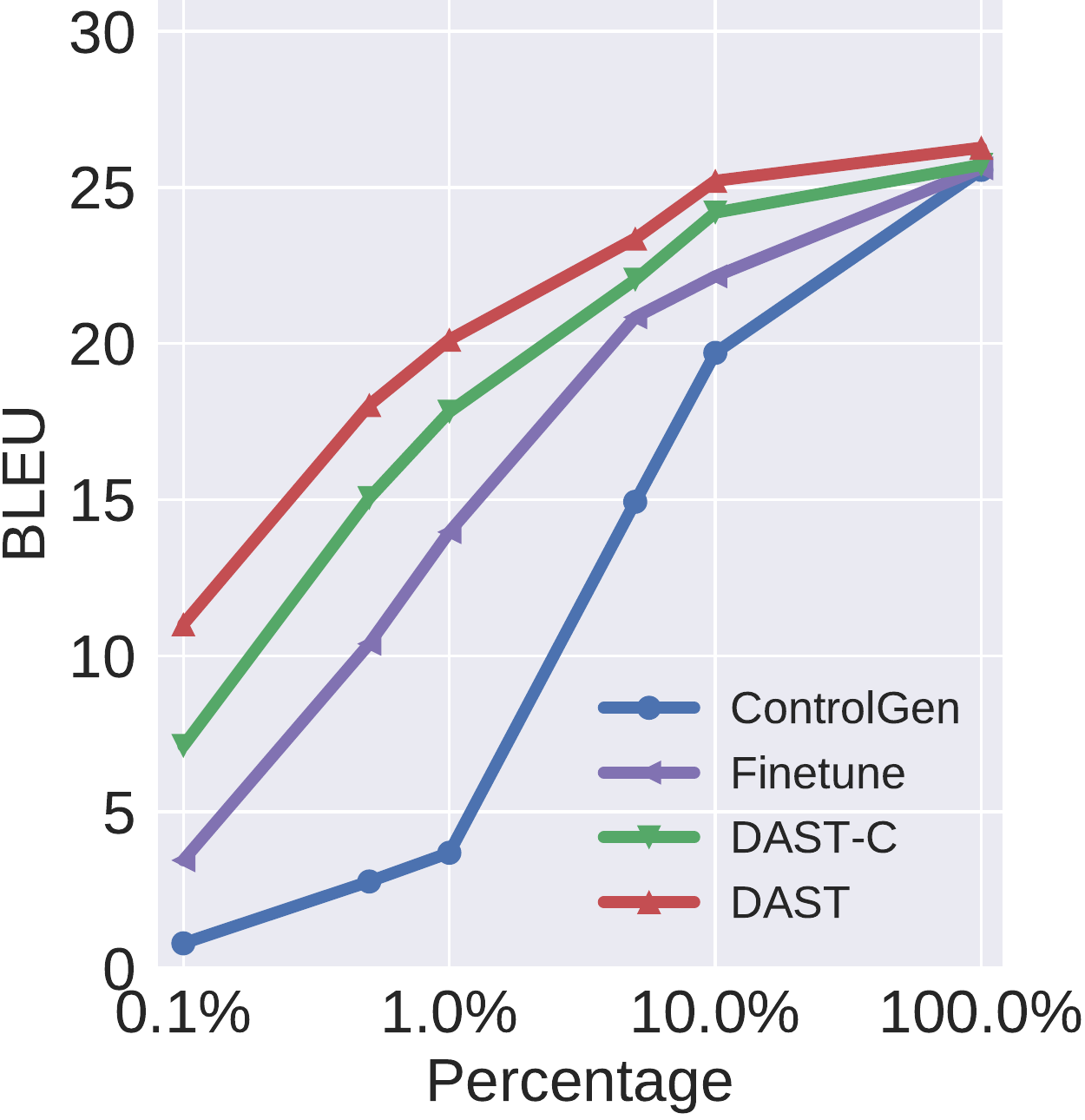}
\endminipage\hfill
\minipage{0.25\textwidth}
  \includegraphics[width=\linewidth]{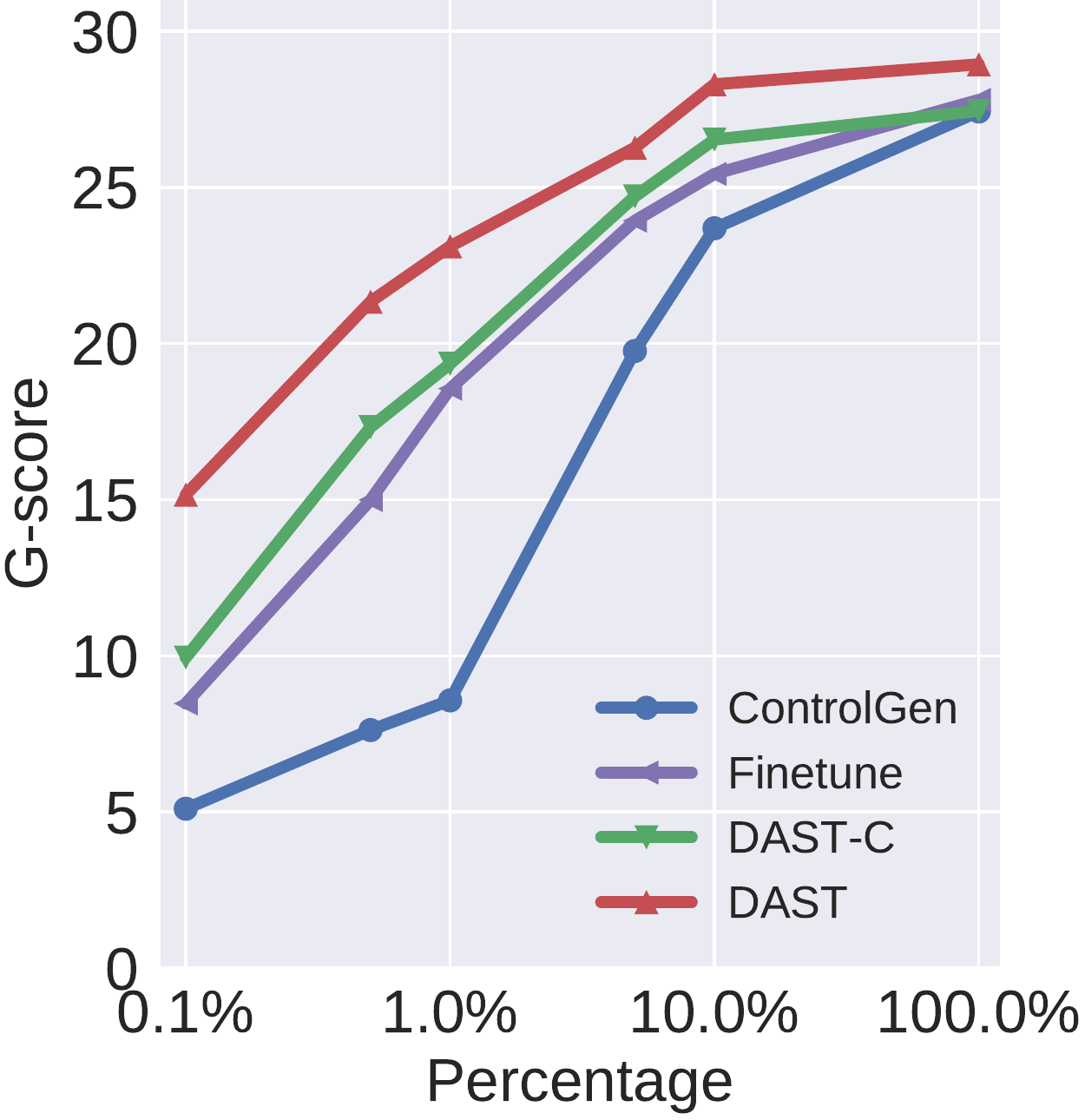}
\endminipage\hfill \\
\caption{Results on Yelp test set in terms of different percentage of target domain data. $0.1\% \approx 400$ samples.}
\label{fig:curves}
\end{figure*}

\begin{table*}[htbp]
\scriptsize
\setlength{\tabcolsep}{3.0pt}
\centering
\begin{tabular}{cc|c|cc c  cc|c|cc c cc|c|cc}
\toprule
\multicolumn{5}{c}{Style Control (Yelp $1\%$ data)} & & \multicolumn{5}{c}{Content Preservation (Yelp $1\%$ data)} & & \multicolumn{5}{c}{Fluency (Yelp $1\%$ data)}\\\cmidrule[\heavyrulewidth]{1-5} \cmidrule[\heavyrulewidth]{7-11} \cmidrule[\heavyrulewidth]{13-17}
\multicolumn{2}{c|}{Our Model} & Neutral & \multicolumn{2}{c}{Comparison} & & \multicolumn{2}{c|}{Our Model} & Neutral & \multicolumn{2}{c}{Comparison} & & \multicolumn{2}{c|}{Our Model} & Neutral & \multicolumn{2}{c}{Comparison}\\\cmidrule[\heavyrulewidth]{1-5} \cmidrule[\heavyrulewidth]{7-11} \cmidrule[\heavyrulewidth]{13-17}
DAST & {\bf56.2\%} & 30.5\% & 13.3\% & ControlGen & &
DAST & {\bf47.0\%} & 48.4\% & 4.6\% & ControlGen & &
DAST & {\bf47.1\%} & 40.8\% & 12.0\% & ControlGen\\

DAST & {\bf40.5\%} & 42.3\% & 17.2\% & DAST-C & &
DAST & {\bf22.4\%} & 65.7\% & 11.9\% & DAST-C & &
DAST & {\bf29.1\%} & 55.8\% & 15.1\% & DAST-C\\

DAST & 17.9\% & 18.5\% & {\bf63.6\%} & Human & &
DAST & 17.7\% & 47.4\% & {\bf34.9\%} & Human & &
DAST & 10.1\% & 30.4\% & {\bf59.5\%} & Human\\\bottomrule

\toprule
\multicolumn{5}{c}{Style Control (Enron)} & & \multicolumn{5}{c}{Content Preservation (Enron)} & & \multicolumn{5}{c}{Fluency (Enron)}\\\cmidrule[\heavyrulewidth]{1-5} \cmidrule[\heavyrulewidth]{7-11} \cmidrule[\heavyrulewidth]{13-17}
\multicolumn{2}{c|}{Our Model} & Neutral & \multicolumn{2}{c}{Comparison} & & \multicolumn{2}{c|}{Our Model} & Neutral & \multicolumn{2}{c}{Comparison} & & \multicolumn{2}{c|}{Our Model} & Neutral & \multicolumn{2}{c}{Comparison}\\\cmidrule[\heavyrulewidth]{1-5} \cmidrule[\heavyrulewidth]{7-11} \cmidrule[\heavyrulewidth]{13-17}
DAST & {\bf74.2\%} & 19.8\% & 6\% & ControlGen & &
DAST & {\bf80.8\%} & 14.8\% & 4.4\% & ControlGen & &
DAST & {\bf73.8\%} & 20.6\% & 5.6\% & ControlGen\\

DAST & {\bf28.4\%} & 50.2\% & 21.4\% & DAST-C & &
DAST & {\bf26.8\%} & 48.8\% & 24.4\% & DAST-C & &
DAST & {\bf26.9\%} & 51.6\% & 21.5\% & DAST-C\\

DAST & 17.6\% & 30.5\% & {\bf51.9\%} & Human & &
DAST & 15.3\% & 36.9\% & {\bf47.8\%} & Human & &
DAST & 11.6\% & 36.5\% & {\bf51.9\%} & Human\\\bottomrule
\end{tabular}
\caption{Results of \textbf{Human Evaluation}
 for style control, content preservation and fluency, showing preferences ($\%$) for DAST model vis-a-vis baseline or other comparison systems. Evaluation results of the \textbf{overall transfer quality} are provided in Appendix~\ref{supp:human_eval}.}
 \label{tab:human}
\end{table*}

\begin{table}[t]
\centering
\small{
\begin{tabular}{c ccc }
\Xhline{3\arrayrulewidth}
\multicolumn{4}{c}{Yahoo Sentiment Transfer} \\\hline
Model  &  D-acc & S-acc   & BLEU \\\hline
ControlGen &- & 99.1 & 9.7\\
Finetune & \textbf{97.8} &98.8 &31.4\\
DAST-C & 90.7 & 98.8 & 35.9\\
DAST & 90.8 & \textbf{99.2} & \textbf{39.2}\\\Xhline{3\arrayrulewidth}
\end{tabular}
}
\caption{Results on Yahoo sentiment transfer task.}
\label{tab:yahoo}
\end{table}

\paragraph{Model Comparisons.}
To evaluate the effectiveness of leveraging massive data from other domains, we compare our proposed DAST models with previously proposed models trained on the target domain (Table~\ref{tab:res}). 
%We also experiment another domain adaptation baseline which is trained on the IMDB and fine-tuned on the target domain (Appendix~\ref{supp:fine_tune}). However, such approach experiences severe catastrophic forgetting~\cite{goodfellow2013empirical} to the source domain information. As a result, the performance is similar as the baseline trained on the target domain data. 
We observe that by leveraging massive data from the IMDB dataset, our models achieve better performance against all baselines on the sentiment transfer tasks in both the Yelp and Amazon domains. 

Notably, when the target domain has limited data ($1\%$), all baselines trained on the target domian only completely fail on content preservation. Finetune preserves better content but experiences the catastrophic forgetting problem~\cite{goodfellow2013empirical} to the source domain information. As a result, the overall style transfer performance is still nonoptimal. On the contrary, with the help of the source domain, DAST obtains considerable content preservation performance improvement when compared with other baselines. Our model also attains favorable performance in terms of style transferring accuracy (S-acc), resulting in a good overall G-score. In general, we observe that DAST-C is able to better preserve content information, while DAST further improves both content preservation and style control abilities. Additionally, both DAST-C and DAST can adapt to the target domain, evidenced by the high domain accuracy (D-acc). The human evaluation results (Table~\ref{tab:human}) show a strong preference of DAST over DAST-C as well as ControlGen in terms of style control, content preservation and fluency.

Finally, we evaluate our models on Yahoo sentiment transfer task. As can be seen in Table~\ref{tab:yahoo}, both DAST and DAST-C achieve successful style transfer even if the target data is formed as questions which have a large discrepancy with the source IMDB domain. The samples of Yelp and Yahoo sentiment transfer are shown in Table~\ref{tab:quali}. We also investigate the effect of different source domain data, included in Appendix~\ref{supp:source_domain}.

\begin{table*}[t]

\centering
\scriptsize
\begin{tabular}{c|l|l}
\hline
{} & Yelp (positive-to-negative) & Yelp (negative-to-positive)\\\hline
\textbf{Input}
& the service was great , food delicious , and the value impeccable . & and the pizza was cold , greasy , and generally quite awful . \\
\textbf{ControlGen}
& the service was \textcolor{red}{horrible} , \textcolor{blue}{service} , \textcolor{orange}{the service and very frustrated} . & and the \textcolor{blue}{food} was \textcolor{red}{delicious}, delicious , and freaking tasty , \textcolor{orange}{delicious} .\\
\textbf{Finetune}
& the service was \textcolor{red}{poor} , food \textcolor{blue}{...} , and the \textcolor{blue}{experience were} . & and the pizza was professional , friendly , and always have \textcolor{red}{great} .\\
\textbf{DAST-C}
& the service was \textcolor{red}{horrible} , food \textcolor{red}{horrible} , and the \textcolor{blue}{slow} sparse . &and the pizza was \textcolor{red}{fresh}, greasy , and generally \textcolor{red}{quite cool} . \\
\textbf{DAST}
& the service was \textcolor{red}{horrible} , food \textcolor{red}{bland} , and the value \textcolor{red}{lousy} . &and the pizza was \textcolor{red}{tasty} , \textcolor{red}{juicy} , and definitely \textcolor{red}{quite amazing} . \\
\textbf{Human}
& service was poor and the food expensive and weak tasting . & the pizza was warm , not greasy , and generally tasted great . \\\hline
{} & Yahoo (positive-to-negative) & Yahoo (negative-to-positive)\\\hline
\textbf{Input}
&who is more romantic ? man or woman ?
  &why do stupid questions constantly receive intelligent answers ?\\
\textbf{ControlGen}
 &\textcolor{blue}{which} is more stupid ? \textcolor{blue}{and or why} ? & \textcolor{blue}{men} do fantastic questions constantly receive intelligent \textcolor{blue}{bound} !\\
 \textbf{Finetune}
 & \textcolor{orange}{the} is more \textcolor{blue}{expensive} ? man or woman ? & why do \textcolor{red}{great} questions \textcolor{blue}{read more entertaining} answers ?\\
\textbf{DAST-C}
&who is more \textcolor{red}{ugly} ? man or woman ?
  & why do \textcolor{red}{important} questions constantly receive intelligent answers ?
 \\
\textbf{DAST}
&who is more \textcolor{red}{crazy} ? man or woman ?
  & why do \textcolor{red}{nice} questions constantly receive intelligent answers ?\\\hline
{} & Enron (informal-to-formal) & Enron (informal-to-formal)\\\hline
\textbf{Input}
&ya 'll need to come visit us in austin .  &are n't you suppose to be teaching some kids or something ?  \\
\textbf{ControlGen}
&\textcolor{blue}{could we need to look on saturday in enpower .} &\textcolor{red}{are you not} supposed to be \textcolor{blue}{disloyal} some kids or something ?  \\
\textbf{Finetune}
& \textcolor{red}{you will} need to go \textcolor{blue}{in bed with him} . &\textcolor{red}{are you not} \textcolor{blue}{to be able to be some man} or something ?  \\
\textbf{DAST-C}
&\textcolor{red}{you will} need to visit \textcolor{blue}{town} .  &are not you supposed to be teaching some kids or something ? \\
\textbf{DAST}
&\textcolor{red}{yes , you will} need to visit us in austin .  &\textcolor{red}{are you not} supposed to be teaching some \textcolor{red}{children} or something ? \\
\textbf{Human}
&all of you should come visit us in austin .  &are you not supposed to be instructing children ?  \\\hline
\end{tabular}
\caption{Transferred sentences on Yelp ($1\%$ data), Yahoo and Enron datasets, where \textcolor{red}{red} denotes successful style transfers, \textcolor{blue}{blue} denotes content losses, and \textcolor{orange}{orange} denotes grammar errors. Better looked in color.}
\label{tab:quali}
\end{table*}

\begin{table}[t]
\centering
\small{
\begin{tabular}{c@{\hskip 1mm}cccc}
\Xhline{3\arrayrulewidth}
Model & D-acc & S-acc & \textit{h}BLEU & G-score\\\hline
DAST & \textbf{97.0} & \textbf{92.6} & \textbf{20.1} & \textbf{23.1} \\\hline
w/o d-spec attributes & 83.9 & 90.9 &20.0 &22.7\\
w/o d-spec classifiers & 91.4 & 83.8 & 19.0 & 20.8\\
w/o both & 73.8 & 80.6 & 18.7 & 19.9\\\Xhline{3\arrayrulewidth}
Setup & D-acc & S-acc & \textit{h}BLEU & G-score\\\hline
IMDB+Yelp & 97.0 & 92.6 & 20.1& 23.1\\
Finetune & 98.1 & 96.7 & 13.9 & 18.5\\
IMDB & 62.8 & 59.3 & 21.4 & 12.2\\
Yelp & 96.8 & 98.5 & 3.7 & 8.6\\\Xhline{3\arrayrulewidth}
%Yelp (finetune) & 98.5 & 98.0 & 8.1 & 12.7\\\bottomrule
\end{tabular}
}
\caption{Ablation study on Yelp (1\%) dataset with help from IMDB dataset. The results are evaluated on Yelp test set. \textit{d-spec} is short for \textit{domain-specific}.}
\label{tab:abl}
\end{table}

\paragraph{Limiting the Target Domain Data.}
We further test the limit of our model by using as few target domain data as possible. 
Figure~\ref{fig:curves} shows the quantitative results with different percentages of target domain training data. When the target domain data is insufficient, especially less than $10\%$, the content preservation ability of the baseline (trained with target data only) has degenerated rapidly despite a relatively high style transfer accuracy. This is not desirable because a transferred sentence can easily have the correct style while barely contains any similar content to the input by retrieving sentences with the target style. Finetune improves content preservation but still suffers the same problem with fewer target data. Note that DAST-C is not comparable to Finetune as the previous one does not use the style information in the source domain.

On the other hand, both DAST models bring substantial improvements to content preservation, and can still successfully manipulate the styles, resulting in consistently higher G-scores. This is presumably because our models adapt the content information as well as the style information from the source domain to consistently sustain the style transfer on the target domain. By learning both generic and domain-specific stylized information, DAST outperforms DAST-C in terms of content preservation and style control. Even with $0.1\%$ target domain data (400 samples), DAST could still attain reasonable text style transfer, whereas the model trained on the target data completely generates sentences in nonsense. Meanwhile, DAST could keep transferring the sentences in a domain-aware manner, achieving high domain accuracy all the time.

\paragraph{Ablation Study.}
To investigate the effect of individual components and training setup on the overall performance, we conduct an ablation study in Table~\ref{tab:abl}. The domain vectors enable the model to transfer sentences in a domain-aware manner, and thus give the largest boost on domain accuracy. Without domain-specific style classifiers, the model mixes the style information on both domains, resulting in worse style control and content preservation. Additionally, simply increasing the number of training data (i.e., the row ``w/o both'') improves content preserving, while introducing a data distribution discrepancy between the training (Yelp+IMDB) and test data (Yelp), as evidenced by the lower S-acc and D-acc scores. 

In terms of the training setup, the source domain IMDB mostly helps content preservation, while accurate style information is mainly learned from the target domain Yelp. Finetune gives higher S-acc and D-acc and lower \textit{h}BLEU due to the catastrophic forgetting. Our proposed DAST uses the source domain data more wisely thus gives balanced results on the style and domain control as well as content preservation.

\begin{table}[t]
\centering
\small{
\begin{tabular}{c  ccc}
\Xhline{3\arrayrulewidth}
\multicolumn{4}{c}{Enron Formality Transfer}\\\hline
Model  &  D-acc & S-acc   & \emph{h}BLEU \\\hline
ControlGen  &- & 81.2 & 4.74\\
Finetune &\textbf{91.3} &81.6 &14.7\\
DAST-C  & 87.6 & 89.2 & 15.5\\
DAST & 88.4 & \textbf{91.6} & \textbf{16.4}\\\Xhline{3\arrayrulewidth}
\end{tabular}
}
\caption{Results on Enron formality transfer tasks.}
\label{tab:enron}
\end{table}

\paragraph{Non-parallel Style Transfer with Parallel Source Data.}
Finally, to verify the versatility of our proposed models on different scenarios, we investigate another domain adaptation setting, where the source domain data (GYAFC) is parallel but the target domain data (Enron) is non-parallel, on the challenging formality transfer task. Since parallel data is available in the source domain, we can simply add a sequence-to-sequence loss $L^{\mathcal{S}}_{s2s}$ on source domain data in Eqn.~\eqref{eq:6} and Eqn.~\eqref{eq:9} to help the target domain without parallel data. The training objectives can be written as: 
$L_{ae}^\mathcal{T}+L_{style}^\mathcal{T}+L_{ae}^{\mathcal{S}}+L_{s2s}^{\mathcal{S}}$ and $L_{ae}^{\mathcal{S,T}} +L_{style}^{\mathcal{S,T}}+L_{s2s}^{\mathcal{S}}$, respectively. Results are summarized in Table~\ref{tab:enron}. DAST outperforms other methods on both style control and content preservation while keeping the transferred sentences with target-specific characteristics (D-acc). A strong human preference for DAST can be observed in Table~\ref{tab:human} when compared to the baselines. Qualitative samples are provided in Table~\ref{tab:quali}.

\section{Conclusion}
We present two simple yet effective domain adaptive text style transfer models that leverage massively available data from other domains to facilitate the transfer task in the target domain. The proposed models achieve better content preservation with the generic information learned from the source domain and simultaneously distinguish the domain-specific information, which enables the models to transfer text in a domain-adaptive manner. Extensive experiments demonstrate the robustness and applicability on various scenarios where the target data is limited.

\section*{Acknowledgments}
We would like to thank the reviewers for their constructive comments. We thank NVIDIA Corporation for the donation of the GPU used for this research. We also thank Hao Peng, Tianyi Zhou for their helpful discussions. 

\bibliography{emnlp-ijcnlp-2019}
\bibliographystyle{acl_natbib}
%\newpage\null\thispagestyle{empty}\newpage
\clearpage
\newpage
\appendix

\section{Supplementary Material}

\subsection{Evaluation Classifiers}
\label{supp:classifers}
We train the style classifier to classify the styles on the target domain. The domain classifiers are trained to distinguish the samples from different domains. After training, all classifiers are used for evaluation only. The test accuracy of evaluation classifiers are reported in Table~\ref{tab:eval_classifer}.
\begin{table}[!ht]
\setlength{\tabcolsep}{3.0pt}
\centering
\small{
\begin{tabular}{cc|cc}
\toprule
\multicolumn{2}{c}{Style Classifier} & \multicolumn{2}{c}{Domain Classifier} \\\midrule
Dataset & Accuracy & Dataset & Accuracy  \\\hline
Yelp & 97.6\% & IMDB \& Yelp & 94.8\%\\\hline
Amazon & 81.0\%& IMDB \& Amazon & 97.1\%\\\hline
Yahoo & 99.4\%& IMDB \& Yahoo & 86.9\%\\\hline
ENRON & 87.0\%& GYAFC \& ENRON & 89.7\%\\
\bottomrule
\end{tabular}
}
\caption{Test accuracy of evaluation classifiers.}
\label{tab:eval_classifer}
\end{table}

\begin{comment}
\subsection{Comparison with Fine-Tuning Method}
\label{supp:fine_tune}
We experiment a simple domain adaptation baseline which trained on the source domain first and then fine-tuned on the target domain. We refer this baseline as \emph{finetune}. As can be seen in~Table~\ref{tab:finetune}, it experiences a severe catastrophic forgetting~\cite{goodfellow2013empirical} to the source domain information when compared with our proposed models. Due to this phenomenon, the \emph{finetune} method converges to a similar performance as the model trained only on the target domain. We hypothesize that this is due to a lack of accurate training signals during the fine-tuning process in the unsupervised text style transfer tasks.

\begin{table}[!ht]
\setlength{\tabcolsep}{5.0pt}
\centering
\small{
\begin{tabular}{c|ccc}
\toprule
Model & D-acc & S-acc & BLEU\\\midrule
\emph{finetune} & 98.5 & 98.0 & 8.1 \\
DAST-C & 96.9 & 90.3 & 17.8 \\
DAST & 97.0 & 92.6 &20.1 \\\bottomrule
\end{tabular}
}
\caption{Performance comparisons with \emph{finetune} model on the Yelp (1\% data).}
\label{tab:finetune}
\end{table}
\end{comment}

\subsection{Source Domain Data}
\label{supp:source_domain}
To investigate the effectiveness of the source domain data, we evaluate our proposed models on different source domains that have unknown styles or the same styles as Yelp. Results are included in Table~\ref{tab:source_domain}. It can be seen that the proposed models can robustly achieve favorable style transfer with help of different source domain data. Since DAST-C model mainly learns the generic content information by modeling the large corpus on the source domain, the number of source training data significantly affects the performance, especially on content preservation (BLEU). On the other hand, DAST also adapts the generic style information, the source domain with closer sentiment information (IMDB) can thus benefit more to the target domain (Yelp) comparing to the TripAdvisor dataset.
\begin{table}[!ht]
\setlength{\tabcolsep}{3.0pt}
\centering
\small{
\begin{tabular}{c|cc|ccc}
\toprule
Model & Source & \# samples& D-acc & S-acc & BLEU\\\midrule
\multirow{3}{*}{DAST-C} & IMDB & 572K & 96.9 & 90.3 & 17.8\\
{} & Yahoo & 900k & 90.3 & 91.3 & 19.6 \\
{} & GYAFC & 206k & 93.5 & 92.9 & 16.1\\\midrule
\multirow{2}{*}{DAST} & IMDB & 334k & 97.0 & 92.6 & 20.1 \\
& TripAdvisor & 572k & 86.2 & 91.4 & 18.4 \\\bottomrule
\end{tabular}
}
\caption{Performance on the Yelp (1\% data) dataset with help of different source domain data.}
\label{tab:source_domain}
\end{table}

\subsection{Human Evaluation}
\label{supp:human_eval}
\begin{figure*}[t]
\centering
\includegraphics[width=1.0\linewidth]{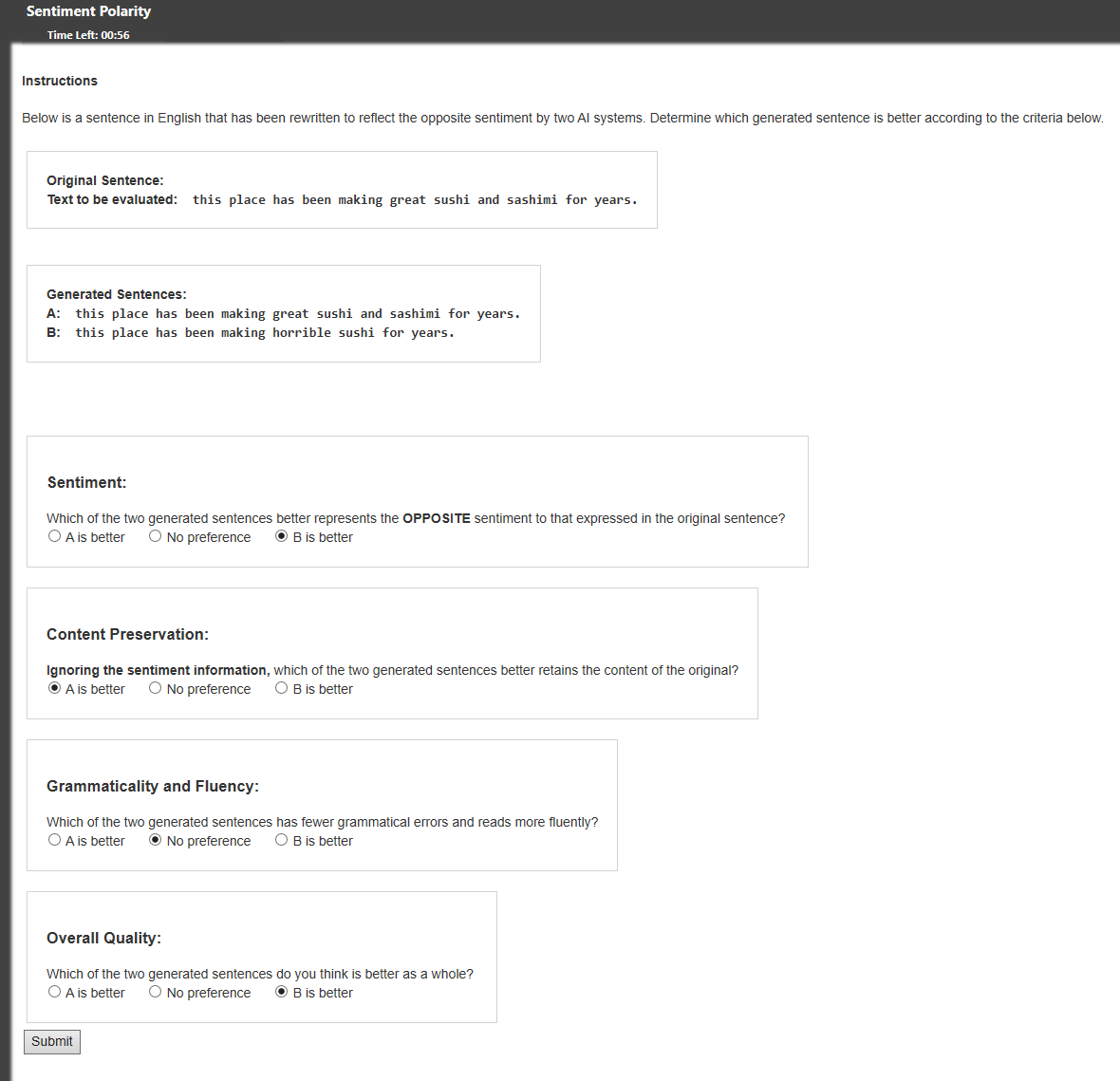}
\caption{Questionnaire used to elicit pairwise judgments from crowd-sourced annotators. Candidate responses were presented in random order.}
\label{fig:questionnaire}
\end{figure*}

For each human evaluation on Yelp sentiment transfer and Enron formality transfer tasks, we randomly sampled 100 sentences from the corresponding test set and collected three responses for each pair on every evaluation aspect, yielding 2700 responses in total. Each pair of system outputs was randomly presented to 7 crowd-sourced judges, who indicated their preference for style control, content preservation and fluency using the form shown in Figure~\ref{fig:questionnaire}. To minimize the impact of spamming, we employed the top-ranked 30\% of U.S. workers provided by the crowd-sourcing service. In order to make the task less abstract, following~\citet{mir2019evaluating}, we asked the judges to evaluate the content preservation quality independently of style information. Detailed task descriptions and examples were also provided to guide the judges. Inter-rater agreement, as measured by agreement with the most common judgment was 75.9\%.

Besides the style control, content preservation and fluency evaluated in Table~\ref{tab:human}, we also asked each worker to provide a judgement of \textbf{the overall quality} in terms of three aspects as a whole. Results are summarized in Table~\ref{tab:overall}. It shows that our DAST model is better in the overall quality compared to the baselines. 

\begin{table}[!ht]
\centering
\small{
\begin{tabular}{cc|c|cc}
\toprule
\multicolumn{5}{c}{Overall Quality (Yelp 1\% data)}\\\midrule
\multicolumn{2}{c}{Our Model} & Neutral & \multicolumn{2}{c}{Comparison}\\\midrule
DAST & \textbf{81.1\%} & 14.0\% & 4.9\% & ControlGen \\
DAST & \textbf{31.4\%} & 43.0\% & 25.6\% & DAST-C \\
DAST & 16.9\% & 23.9\% & \textbf{59.2\%} & human \\\bottomrule\toprule

\multicolumn{5}{c}{Overall Quality (Enron)}\\\midrule
\multicolumn{2}{c}{Our Model} & Neutral & \multicolumn{2}{c}{Comparison}\\\midrule
DAST & \textbf{52.7\%} & 35.3\% & 12.0\% & ControlGen \\
DAST & \textbf{34.0\%} & 48.4\% & 17.6\% & DAST-C \\
DAST & 12.0\% & 17.8\% & \textbf{68.0\%} & human \\\bottomrule
\end{tabular}
}
\caption{ Results of \textbf{Human Evaluation} in terms of the overall quality on Yelp sentiment transfer and Enron formality transfer tasks.}
\label{tab:overall}
\end{table}

\end{document}